\documentclass[12pt,twoside]{article}
\usepackage{booktabs}
\usepackage{array} 
\usepackage{graphicx} 
\usepackage{subfig}
\usepackage{booktabs}   
\usepackage{caption}
\usepackage{tabularx}
\usepackage{graphicx}
\usepackage{tabularx}
\usepackage{multirow}
\usepackage{subcaption} 
\usepackage{graphicx}%
\usepackage{multirow}%
\usepackage{amsmath,amssymb,amsfonts}%
\usepackage{amsthm}%
\usepackage{mathrsfs}%
\usepackage[title]{appendix}%
\usepackage{xcolor}%
\usepackage{textcomp}%
\usepackage{manyfoot}%
\usepackage{booktabs}%
\usepackage{algorithm}%
\usepackage{algorithmicx}%
\usepackage{algpseudocode}%
\usepackage{listings}%
\usepackage{authblk}
\usepackage{float}
\newcommand{\keywords}[1]{\noindent{\small\textbf{Keywords:} #1}}

\begin{document}
\title{CovidLLM: A Robust Large Language Model with Missing Value Adaptation and Multi-Objective Learning Strategy for Predicting Disease Severity and Clinical Outcomes in COVID-19 Patients}

\author[1]{Shengjun Zhu}
\author[2]{Siyu Liu}
\author[3]{Yang Li}
\author[4]{Qing Lei}
\author[4]{Hongyan Hou}
\author[3]{Hewei Jiang}
\author[3]{Shujuan Guo}
\author[4]{Feng Wang}
\author[2]{Rongshang Chen}
\author[4]{Xionglin Fan}
\author[3]{Shengce Tao}
\author[1]{Jiaxin Cai}

\affil[1]{School of Mathematics and Statistics, Xiamen University of Technology, Xiamen, China \\ (e-mail: 19359446568@163.com; caijiaxin@xmut.edu.cn)}

\affil[2]{School of Computer and Information Engineering, Xiamen University of Technology, Xiamen,  China \\(e-mail: 1114595740@qq.com; 2005120705@xmut.edu.cn)}

\affil[3]{Shanghai Center for Systems Biomedicine, Key Laboratory of Systems Biomedicine (Ministry of Education), Shanghai Jiao Tong University, Shanghai, China \\ (e-mail: liyang202202@163.com; jhw622@sjtu.edu.cn; shjguo@sjtu.edu.cn; taosc@sjtu.edu.cn)}

\affil[4]{Tongji Medical College, Huazhong University of Science and Technology, Wuhan, China \\ (e-mail: leiqinghust@163.com; houhongyan89@163.com; fengwang@tjh.tjmu.edu.cn; xlfan@hust.edu.cn)}



\date{This work is supported by the Natural Science Foundation of Fujian Province (2023J05083, 2022J011396, 2023J011434). \\ Shengjun Zhu, Siyu Liu, and Yang Li are co-first authors. Corresponding authors: Jiaxin Cai and Shengce Tao.}

\maketitle

\begin{abstract}
Coronavirus Disease 2019 (COVID-19), which emerged in 2019, has caused millions of deaths worldwide.
Although effective vaccines have been developed to mitigate severe symptoms, certain populations, particularly the elderly and those with comorbidities, remain at high risk for severe outcomes and increased mortality.
Consequently, early identification of the severity and clinical outcomes of the disease in these patients is vital to prevent adverse prognoses.
Although traditional machine learning and deep learning models have been widely employed in this area, the potential of large language models (LLMs) remains largely unexplored.
Our research focuses primarily on constructing specialized prompts and adopting multi-objective learning strategies. 
We started by selecting serological indicators that significantly correlate with clinical outcomes and disease severity to serve as input data for the model.
Blood test samples often contain numerous missing values, and traditional models generally rely on imputation to handle these gaps in the data. In contrast, LLMs offer the advantage of robust semantic understanding. By setting prompts, we can explicitly inform the model when a feature's value is missing, without the need for imputation. 
For the multi-objective learning strategy, the model is designed to first predict disease severity and then predict clinical outcomes. 
Given that LLMs utilize both the input text and the generated tokens as input for generating the next token, the predicted severity is used as a basis for generating the clinical outcome. 
During the fine-tuning of the LLM, the two objectives influence and improve each other. Our experiments were implemented based on the ChatGLM model. The results demonstrate the effectiveness of LLMs in this task, suggesting promising potential for further development.
All code is available at {https://github.com/sysll/CovidLLM} .
\end{abstract}


\section{Introduction}
COVID-19 emerged in 2019, caused by the severe acute respiratory syndrome coronavirus 2 (SARS-CoV-2), placing immense pressure on the world. Most patients exhibited symptoms such as cough, muscle pain, dizziness, and sore throat \cite{fernandez2021prevalence, weng2021pain, fernandez2021headache, cirulli2020long, mahase2021covid, kahraman2020mucosal}, ranging from severe (e.g., acute respiratory distress syndrome (ARDS) and organ failure \cite{berlin2020severe, he2021clinical}) to mild (e.g., intermittent dizziness and minor cough). 
While the overall impact of COVID-19 has diminished, the virus continues to circulate globally, with new variants emerging. 
This ongoing threat is particularly acute for the elderly and individuals with comorbidities, who are at greater risk of severe illness and adverse outcomes \cite{hu2021cytokine}.
Therefore, timely identification of patients at high risk for severe complications or death is crucial, especially since some effective treatments must be administered early in the disease course \cite{hu2021cytokine}. 
Traditional machine learning and deep learning approaches have been widely applied to predict disease severity and clinical outcomes \cite{alballa2021machine, chieregato2022hybrid, wu2021interpretable, hu2020early, patterson2021immune, sayed2021applying, raman2023machine}, typically by inputting a set of serological indicators to predict patient severity or outcomes. However, few studies have explored the potential application of LLMs in this task. 
The primary distinction between LLMs and traditional models lies in their pre-training on extensive human language datasets, which endows them with language comprehension abilities \cite{huang2023chatgpt}. 
Moreover, LLMs have been extensively applied to various prediction tasks and have shown great prospects. 
The study by Hao Xue et al. \cite{xue2023promptcast} designed time-series data as prompts for LLMs and fine-tuned the LLM to predict future data. They were the first to propose a paradigm that designs data as prompts, transforming the prediction task into a dialogue task. 
Liang et al. \cite{liang2024exploring} combined textual description data with taxi trip data to predict crowd flows, significantly improving the accuracy of predictions during holidays. They suggested that textual data could contribute to prediction results to some extent. 
Ding et al. \cite{ding2024semantic} used LLMs to capture the correlations between data, enabling the imputation of missing data in recommendation systems, and demonstrated that this approach outperformed traditional methods. 
Jin et al. \cite{jin2023time} proposed time series forecasting by reprogramming large language models (Time-LLM), where they reprogrammed time-series data into more natural textual prototypes for LLMs and provided textual guidance within the prompt to assist the model in making predictions. Jin et al. argued that LLMs show great potential for time-series forecasting.

\begin{figure}
\centering 
\includegraphics[width=0.9\textwidth]{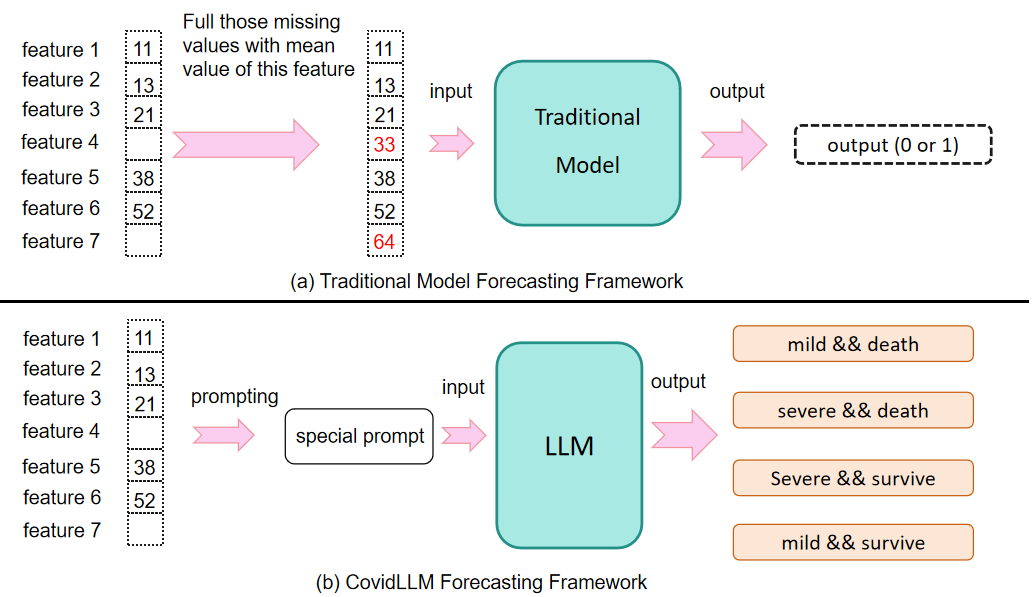}
\caption{(a) The prediction framework of traditional models, which typically involves single-objective prediction and requires imputation of missing values. (b) The prediction framework is based on large language models, which support multi-objective prediction and do not require the imputation of missing values.} 
\label{pic1}
\end{figure}

LLMs also exhibit distinct advantages and potential in our task. 
For serological datasets, which often contain numerous indicators and substantial amounts of missing values, traditional models require the imputation of missing data to meet input format requirements (e.g., using the mean or median of the feature, as shown in Figure \ref{pic1}(a)). However, we believe this approach sacrifices a degree of accuracy to conform to input constraints. 
LLMs, in contrast, do not require such imputation. Due to their powerful language comprehension capabilities, we can simply indicate in the prompt that “this feature’s value is missing.” 
The model can understand this information and direct its attention to other features. If we use the mean or median to fill in missing data, it may mislead the model, which could subsequently affect the prediction results.
Additionally, we propose a multi-objective learning strategy, as shown in Figure \ref{pic1}(b). We think that there is a strong correlation between a patient’s severity and their clinical outcome. 
When physicians assess a patient’s clinical outcome, they usually first determine the severity of the patient’s condition. 
Our multi-objective strategy operates on a similar principle.
LLMs have distinct advantages in multi-objective learning as well. 
Specifically, most large language models predict the next token based on both the model’s input and the previously generated tokens, continuing until the complete output is generated. 
In our multi-objective learning strategy, the model first outputs the severity of the patient and then the clinical outcome. 
This means that the predicted severity informs the subsequent clinical outcome determination. 
Additionally, without specialized model design, traditional models generally can not be used for multi-objective prediction tasks. 
Thus, LLMs have broad potential for exploration in this task and can also be extended to similar tasks. We summarize our research contributions in the following three points.

\begin{itemize} 

\item  We propose using large language models to predict the severity and clinical outcomes of COVID-19 patients and validate the effectiveness of this approach on our collected dataset.

\item  We introduce a novel missing value handling method based on the characteristics of large language models, allowing the model to ignore such features. This improves the model’s robustness to missing data.

\item  Leveraging the autoregressive nature of large language model outputs, we propose a multi-objective learning strategy where the model first predicts the patient's disease severity, followed by their clinical outcome. This approach aligns with standard diagnostic procedures and enhances model performance.

\end{itemize}

\section{Dataset and pre-processing method}
In this study, we systematically collected data from hospitalized COVID-19 patients treated at Tongji Hospital in Wuhan, China, between January and May 2020. Each patient underwent multiple serological tests during their hospitalization. 
Therefore, our dataset includes blood sample data from various time points for each patient, encompassing rich demographic features and detailed records of key serological indicators. Specifically, our dataset includes information from 616 patients, with a total of 6,483 blood test samples collected. The statistical details of this dataset are presented in Table \ref{tableChar}.

\begin{table}[htbp]  
\centering  
\caption{Descriptive statistics for the clinical characteristics of patients diagnosed with COVID-19. \small Continuous variables are presented as means, while categorical variables are shown as counts (percentages).}   

\begin{tabularx}{\textwidth}{XX}\hline
Variable & Case \\  
\midrule  
Age (years) & 61 (14.5) \\  
Severity: &  \\  
\quad Severe & 248 (40.2\%) \\  
\quad Mild & 368 (59.8\%) \\  
Clinical Outcome: &  \\  
\quad Death & 46 (7.4\%) \\  
\quad Survival & 570 (92.6\%) \\  
Sex: &  \\  
\quad Male & 299 (48.5\%) \\  
\quad Female & 317 (51.5\%) \\  
\hline
\end{tabularx}  
\label{tableChar}
\end{table}

\begin{figure}
\centering 
\includegraphics[width=0.8\textwidth]{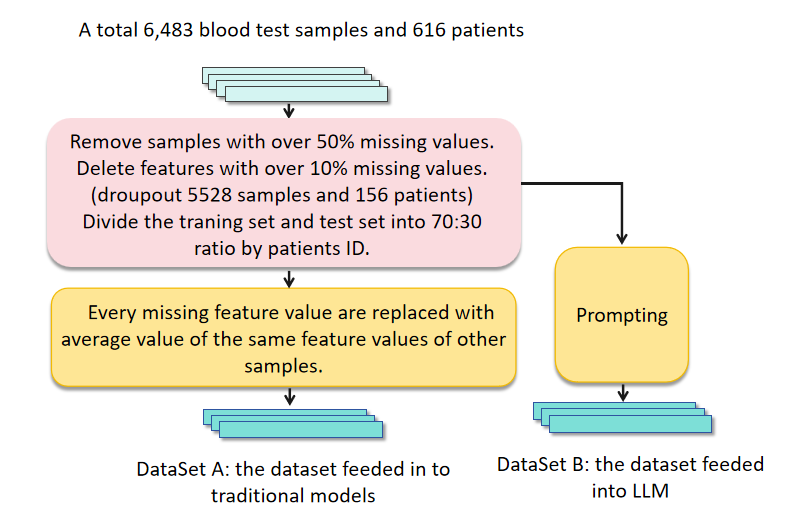}
\caption{Data preprocessing flowchart.} 
\label{pic2} 
\end{figure}

The data preprocessing flowchart is shown in Figure \ref{pic2}. Our dataset is divided by patient, ensuring that no samples from the same patient appear in both the training and testing sets.
The training sets of Dataset A and Dataset B include 322 patients and 679 blood test samples, while the testing sets consist of 138 patients and 276 blood test samples. 
After removing features with more than 10\% missing values, we retained a total of 65 features. 
Those features include Sex, Age, Hypertension (HBP), Diabetes Mellitus (DM), Hypertension (HBP), Diabetes Mellitus (DM),Hemoglobin (HGB), Lymphocyte Percentage (LYMPH\%), Mean Corpuscular Hemoglobin (MCH), Mean Corpuscular Hemoglobin Concentration (MCHC),  Mean Corpuscular Volume(MCV), Absolute Monocyte Count (Mono\#), Monocyte Percentage 
(MONO(\%) ), Absolute Neutrophil Count (Neu\#),  Neutrophil Percentage (Neu\%), Platelet Count (PLT), Red Blood Cell Count (RBC),Absolute Basophil Count (Baso\#), Basophil Percentage (Baso\%), Eosinophil Count (Eos(\#)), Eosinophil Percentage (Eos(\%)), Hematocrit (HCT), White Blood Cell Count (WBC),  Red Cell Distribution Width CV (RDW\_CV), Red Cell Distribution Width SD (RDW\_SD), Mean Platelet Volume (MPV), Platelet Distribution Width (PDW), Platelet -larger cell ratio (P-LCR), Plateletcrit (PCT), Potassium (K), Calcium (Ca), Na (Sodium), Chloride (Cl), Alanine Aminotransferase (ALT), Lactic Dehydrogenase (LDH), Lactic Dehydrogenase multiplied by 0.9 (LDH0.9), Aspartate Aminotransferase (AST), Total Cholesterol (TC), Albumin (ALB), Alkaline Phosphatase (ALP), Gamma-Glutamyl Transferase ($\gamma$-GT), Total Bilirubin (TBIL), Total Protein (TP), Albumin/Globulin Ratio (A/G Ratio), Globulin (GLOB), Total Protein multiplied by 0.75 (TP0.75), Direct Bilirubin (BC), Total Bilirubin multiplied by 0.8 (TBIL0.8), Indirect Bilirubin (IBil), Creatinine (Cre), Urea, Uric Acid (UA), Bicarbonate (HCO3-), Estimated Glomerular Filtration Rate  (eGFR), Calcium Correction (CC), hypersensitive C-reactive protein (hs-CRP), Quantitative D-dimer assay, Prothrombin time (PT), Prothrombin Activityprothrombin Time Activity (PTA), International Normalized Ratio (INR), Activated Partial Thromboplastin Time (APTT), Fibrinogen (FIB), Thrombin Time (TT). 
In Dataset A, we classified severe and mild disease severity as 1 and 0, respectively, and clinical outcomes of death and survival as 1 and 0.

\section{Method}

\subsection{Specific prompt design}
Our model utilizes a specialized prompt design to improve its robustness in processing serum data.
An example of this prompt design is illustrated in Figure \ref{pic3}. 
We will first design an instruction to tell LLM the task background. The instruction is "As an experienced clinical medicine expert, predict COVID-19 severity (severe/mild) and predict clinical outcome (survive/death) based on serum report. The serum report is as follows.". 
Then, all individual feature values in the model input are transformed into a prompt for the LLM.
When a feature value is missing, we explicitly inform the model of its absence by design text "This feature's value is missing". 
The LLMs, which utilize a self-attention architecture, can effectively interpret this semantic information. 
Upon receiving a "feature's value missing" indication, the model will allocate less attention to that feature and redirect its focus to other features. 
As a result, the model predicts the severity of the patient's disease and clinical outcomes based solely on non-missing feature values, thereby effectively eliminating potential disturbances caused by missing feature values and ensuring the accuracy and reliability of the prediction results.
This approach will improve the model's robustness in the presence of missing values.

\subsection{Multi-Objective Learning Strategy}
We begin by highlighting the characteristics of autoregressive models, such as ChatGLM. This model leverages the self-attention mechanism of the Transformer architecture, which enables it to effectively capture long-range dependencies within the input sequence. 
During the autoregressive generation process, it initiates the sequence with an initial symbol and progressively generates subsequent words, with each output intricately dependent on both the input and the content previously generated.

\begin{figure}
\centering 
\includegraphics[width=1\textwidth]{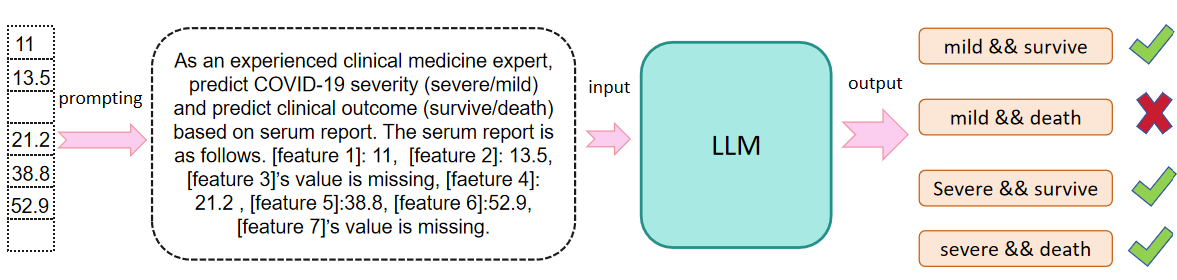}
\caption{The data input and output of the CovidLLM model.} 
\label{pic3} 
\end{figure}

Based on the autoregressive characteristics, we propose our multi-objective strategy. The model first predicts the severity of the patient based on the digital information and text within the prompts we input. Then, the model combines the prediction results of severity and the data information and text within the prompts again to predict the clinical outcome of the patient.
Broadly, the model's outputs fall into four categories, (1) mild and survive, (2) mild and death, (3) severe and survive, and (4) severe and death, as illustrated in Figure \ref{pic3}. 
Through our multi-objective learning strategy, the first severity prediction result informs the clinical outcome prediction. If severity is assessed as mild, the likelihood of death becomes nearly impossible. 
Therefore, the model's output, (2) mild and death, are excluded. 
It greatly reduces the chance of making one type of mistake.

\subsection{Overall Framework}
We first employed the Gradient Boosting algorithm for feature selection on the training set of DataSet A, identifying the top five features most relevant to disease severity and clinical outcomes, respectively. 
We then took the union of these features as input for traditional model predictions, which also served as the foundational data for constructing prompts for the large language model (LLM). 
For traditional models, due to structural limitations, we can only perform single-objective prediction tasks. Therefore, we trained the traditional models on the training set of DataSet A and evaluated it on its test set. 
For our CovidLLM model, we adopted a multi-objective learning strategy to predict two objectives simultaneously. 
Our CovidLLM is implemented based on the ChatGLM model and was trained using the P-tuning fine-tuning method on the training set of DataSet B, with evaluation conducted on its test set.
DataSet A and DataSet B are two forms of a dataset, where DataSet A is processed for traditional models and DataSet B is processed for LLMs.

\subsection{Implement Detail}

\subsubsection{LLM and feature selection model}
Our approach is implemented based on ChatGLM-6b \cite{glm2024chatglm}, an autoregressive dialogue generation model built on the Transformer architecture, which excels in language understanding and generation, facilitating bilingual conversations in Chinese and English.
We employ the P-tuning method for fine-tuning, which utilizes hyperparameters provided in the GitHub repository for this project. 
Unlike traditional methods that adjust prompts solely at the input layer, P-tuning \cite{liu2022p} integrates prompt tokens (i.e., embedding vectors from the Prefix Encoder) at each layer of the Transformer blocks. 
This approach enables the model to respond more effectively to task-specific requirements and enhances its predictive capabilities at deeper levels.
For feature selection, we utilize the Gradient Boosting algorithm. 
During training, this algorithm assesses the importance of features based on their contribution to the splits in each decision tree. 
This mechanism allows it to automatically identify which features significantly enhance the predictive performance of the model.
Our parameter settings include n\_estimators = 100, learning rate = 0.01, and max depth = 3.
All experiments are implemented in Python 3.6 with TensorFlow 1.14.0 on a computer with CPU Intel Xeon Gold 6138 @ 2.00GHz (40 cores) and GPU NVIDIA RTX2080Ti. 
All code is available at {https://github.com/sysll/CovidLLM} .

\subsubsection{Compared models}
Adaptive Boosting (AdaBoost)\cite{AdamBoost} is a widely used ensemble learning algorithm introduced by Yoav Freund in 1995. 
Its primary aim is to improve the accuracy and generalization of the classification by combining multiple weak classifiers, such as decision trees, into a robust classifier.
The Gradient Boosting model\cite{gradientboost} builds on this concept by iteratively training multiple weak learners and aggregating their output to create a strong learner. 
This approach systematically refines the predictions of the model through each iteration.
Random Forest\cite{ho1998random} is another ensemble learning method that constructs numerous decision trees based on the decision tree algorithm. 
It synthesizes the predictions from these trees to produce a final outcome, improving overall accuracy.
The K nearest neighbor classifiers (KNN)\cite{peterson2009k} operate on the principle of measuring the distances between the data points. 
This algorithm identifies the K nearest neighbors to the predicted point and determines its class or value based on the characteristics of these neighbors.

\section{Result}
\subsection{Metrics}
We evaluated our model using several key metrics, precision, recall, F1 score, and accuracy (ACC). Precision measures the reliability of the model's positive predictions, indicating the proportion of true positives among all predicted positives. Recall focuses on the model's ability to identify all actual positive samples. 
The F1 score serves as the harmonic mean of precision and recall, providing a balanced assessment of model performance. 
Lastly, accuracy represents the overall proportion of correct predictions made by the model, serving as a straightforward indicator of its effectiveness.
The details of those formulas are as follows.

\begin{equation}
\text{Precision} = \frac{TP}{TP + FP}
\end{equation}

\begin{equation}
\text{Recall} = \frac{TP}{TP + FN}
\end{equation}

\begin{equation}
F1\text{ Score} = 2 \cdot \frac{\text{Precision} \cdot \text{Recall}}{\text{Precision} + \text{Recall}}
\end{equation}

\begin{equation}
\text{Accuracy} = \frac{TP + TN}{TP + TN + FP + FN}
\end{equation}

TP (True Positive) denotes the number of samples correctly predicted as positive.
FP (False Positive) denotes the number of samples incorrectly predicted as positive.
TN (True Negative) denotes the number of samples correctly predicted as negative.
FN (False Negative) denotes the number of samples incorrectly predicted as negative. We think the ACC and F1 scores are the best metrics.

\subsection{Feature selection result}
\begin{figure}
\centering 
\includegraphics[width=0.9\textwidth]{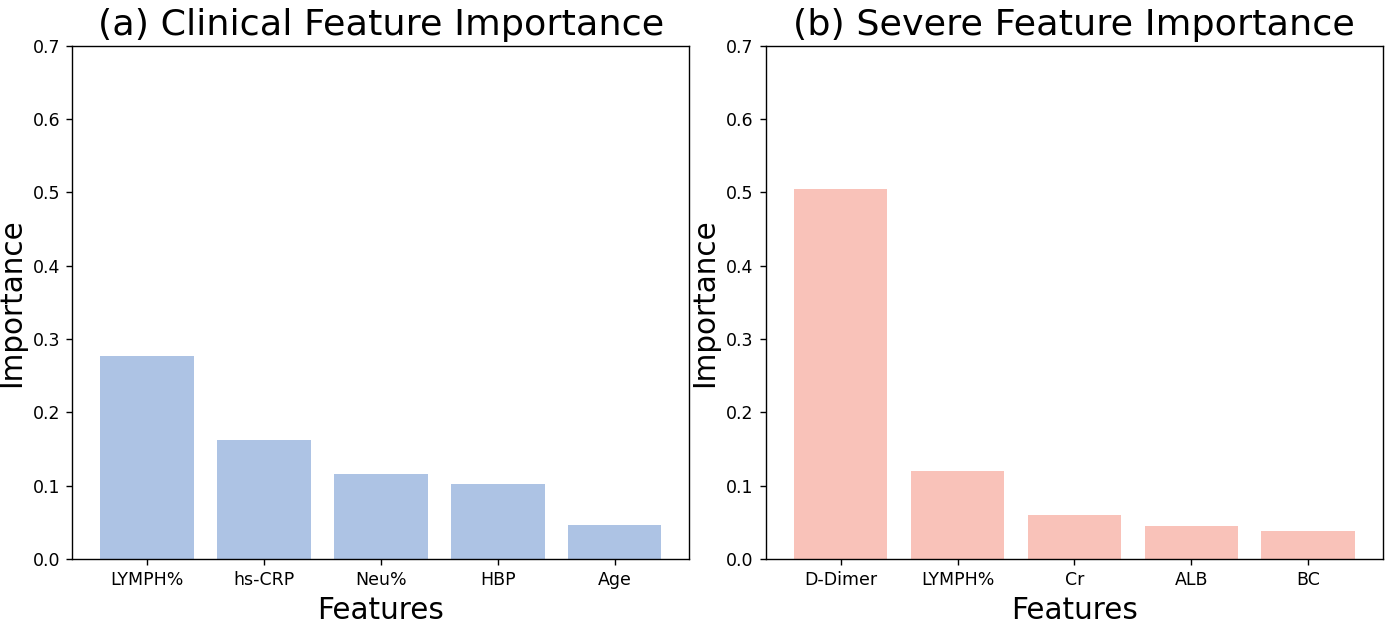}
\caption{(a) is the features related to severity and the corresponding importance weights. (b) is the features related to clinical outcomes and the corresponding importance values.} 
\label{pic4} 
\end{figure}

We employed the GradientBoost model to perform feature selection for both disease severity and clinical outcomes. 
We identified the top five features relevant to each objective, along with their corresponding importance values. 
As shown in Figure \ref{pic4}, the top five features associated with clinical outcomes are Lymphocyte Percentage (LYMPH\%), hypersensitive C-reactive protein (hs-CRP), Neutrophil Percentage (Neu\%), Hypertension (HBP), and Age.
For disease severity, the top five features include D-Dimer, Lymphocyte Percentage (LYMPH\%), Creatinine (Cre), Albumin (ALB), and Indirect Bilirubin (BC). 
Then, we took the union of these two feature sets. This resulting set was utilized in traditional models to predict disease severity and clinical outcomes, and it also served as foundational data for constructing prompts for the large language model.
The union of these features includes LYMPH\%, Age, hs-CRP, Neu\%, HBP, D-Dimer, Cre, ALB, and BC.

\subsection{Prediction of severity}
\begin{figure}
\centering 
 \includegraphics[width=0.9\textwidth]{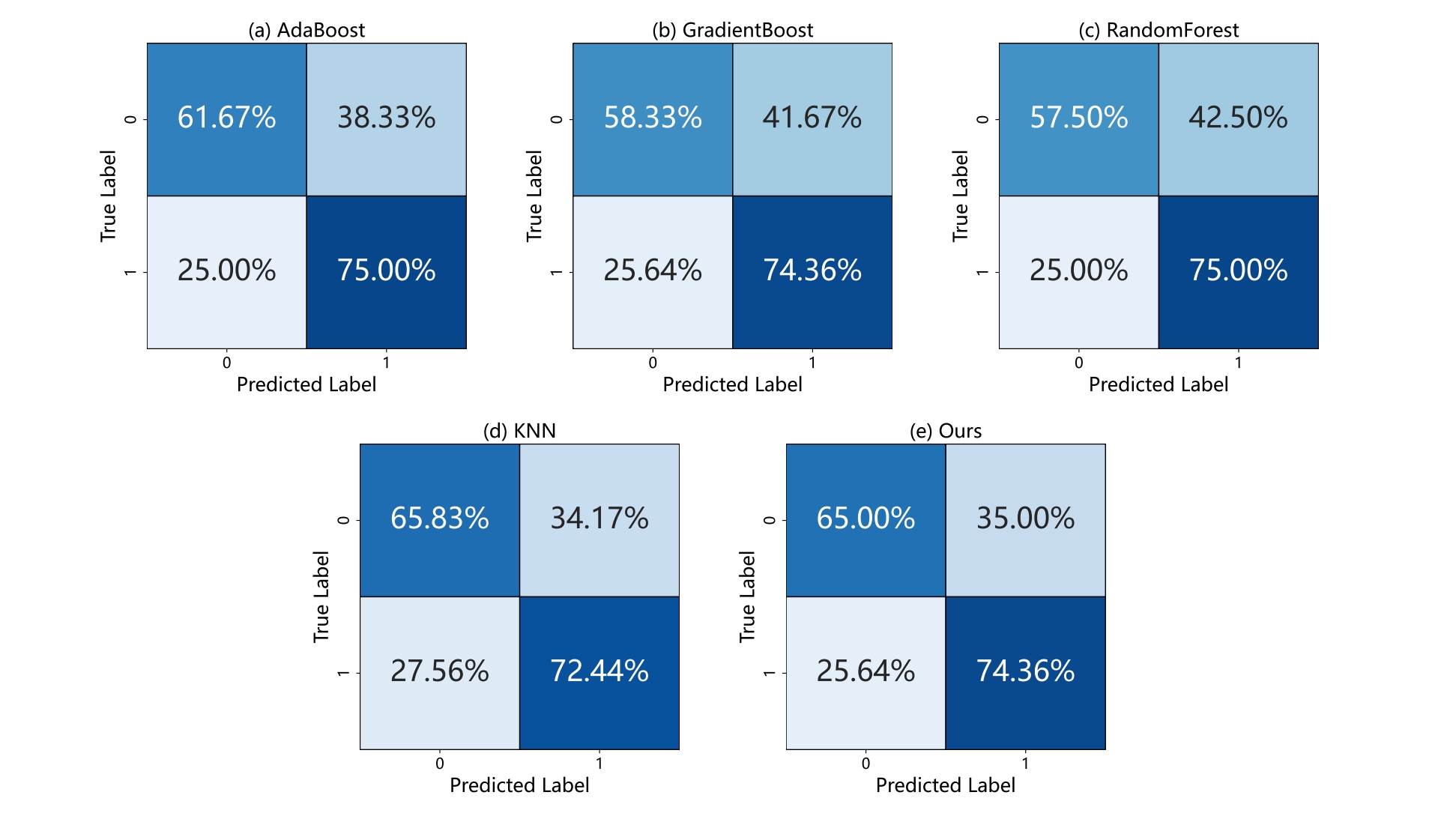}
\caption{Confusion matrices for the CovidLLM model and comparative models. (a) - (d) represents the comparative models, while (e) depicts the CovidLLM.} 
\label{pic5} 
\end{figure}

Figure \ref{pic5} presents the confusion matrix for our model compared to several baseline models in the task of predicting disease severity. 
We observed that traditional models such as GradientBoost, AdaBoost, and RandomForest tend to misclassify patients as having severe cases. 
In contrast, our proposed CovidLLM and KNN models do not exhibit this tendency, maintaining strong predictive performance across both categories.
Table \ref{table 2} summarizes the performance comparison between our model and other baseline models. 
We can observe that CovidLLM (our model) can achieve the best ACC of 70.29\% and its performance in predict class 0 and class 1 achieves the best F1-score. Additionally, our model obtained the best precision in two-class prediction.

\begin{table}[ht]
\centering
\caption{The performance comparison between our model and other models in severity prediction task. The maximum value is indicated in bold. }
\begin{tabular}{ccccc}
\toprule
\textbf{Model} & \textbf{Accuracy} & \textbf{Precision} & \textbf{Recall} & \textbf{F1-score} \\
\midrule
AdamBoost& 0.6920 &        &        &        \\
0        &        & 0.6549 & 0.6167 & 0.6352 \\
1        &        & 0.7178 & \textbf{0.7500} & 0.7335 \\

\midrule
GDBT     & 0.6739 &        &        &        \\
0        &        & 0.6364 & 0.5833 & 0.6087 \\
1        &        & 0.6988 & 0.7436 & 0.7205 \\
\midrule
RandomForest & 0.6739 &        &        &        \\
0            &        & 0.6389 & 0.5750 & 0.6053 \\
1            &        & 0.6964 & \textbf{0.7500} & 0.7222 \\
\midrule
KNN      & 0.6956 &        &        &        \\
0        &        & 0.6475 & \textbf{0.6583} & 0.6529 \\
1        &        & 0.7338 & 0.7244 & 0.7290 \\
\midrule
Our         & \textbf{0.7029} &                 &  
            &        \\
0           &        & \textbf{0.6610} & 0.6500 & \textbf{0.6555} \\
1           &        & \textbf{0.7342} & 0.7436 & \textbf{0.7389} \\
\bottomrule
\end{tabular}
\label{table 2}
\end{table}

\subsection{Prediction of clinic outcome}
\begin{figure}
\centering 
\includegraphics[width=0.9\textwidth]{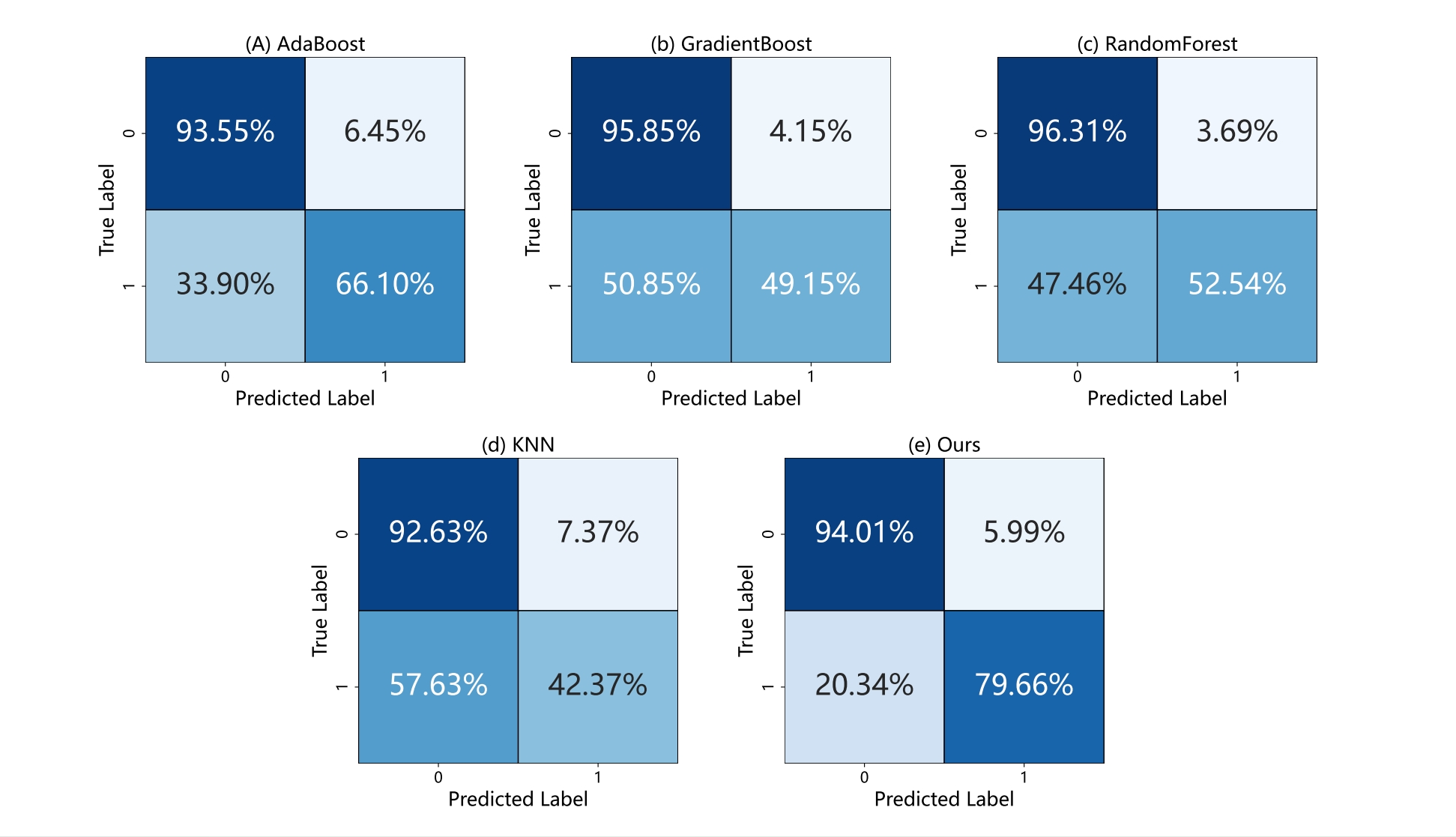}
\caption{Confusion matrices for the CovidLLM model and comparative models. (a) - (d) represents the comparative models, while (e) depicts the CovidLLM.}
\label{pic6}
\end{figure}

Figure \ref{pic6} illustrates the confusion matrix for our proposed model compared to baseline models in predicting clinical outcomes. 
Analysis of this matrix reveals that traditional models tend to misclassify patients as category 0 (survive), leading to an overestimation of accuracy for this category. 
In contrast, our language model-based prediction method maintains high accuracy across both categories, demonstrating greater robustness.
Table \ref{table 3} provides a comparative analysis of our model against baseline models across multiple performance metrics. 
We can observe that our model shows a significant improvement compared to traditional models. Our model achieves an ACC of 90.94\%. Notably, our model's F1 score of class 1 has a great improvement. Additionally, our model's precision of class 0 is 94.44\%, which is close to the clinic application standard.

\begin{table}[ht]
\centering
\caption{The performance comparison between our model and other models in clinic prediction task. The maximum value is indicated in bold. }
\begin{tabular}{ccccc}
\toprule
\textbf{Model} & \textbf{Accuracy} & \textbf{Precision} & \textbf{Recall} & \textbf{F1-score} \\
\midrule
AdamBoost& 0.8768 &        &        &        \\
0        &        & 0.9103 & 0.9355 & 0.9227 \\
1        &        & 0.7358 & 0.6610 & 0.6964 \\

\midrule
GDBT     & 0.8586 &        &        &        \\
0        &        & 0.8739 & 0.9585 & 0.9143 \\
1        &        & 0.7632 & 0.4915 & 0.5979 \\
\midrule
RandomForest & 0.8695 &        &        &        \\
0            &        & 0.8819 & \textbf{0.9631} & 0.9207 \\
1            &        & \textbf{0.7949} & 0.5254 & 0.6327 \\
\midrule
KNN      & 0.8188 &        &        &        \\
0        &        & 0.8553 & 0.9263 & 0.8894 \\
1        &        & 0.6098 & 0.4237 & 0.5000 \\
\midrule
Our         & \textbf{0.9094} &                 &  
            &        \\
0           &        & \textbf{0.9444} & 0.9401 & \textbf{0.9423} \\
1           &        & 0.7833 & \textbf{0.7966} & \textbf{0.7899} \\
\bottomrule
\end{tabular}
\label{table 3}
\end{table}

\section{Discussion}
Our research investigates the application of LLMs in predicting the severity and clinical outcomes of COVID-19 patients, offering new insights into patient prognosis. 
This approach is not limited to COVID-19 but can also be applied to other diseases.
Specifically, we leverage the capability of LLMs to process textual information by directly informing the model when certain feature values are missing. 
This enables the model to focus on available features while effectively ignoring those that are absent. 
We also propose a multi-objective learning strategy that capitalizes on the autoregressive generation capabilities of the model.
In this framework, the model first predicts patient severity and subsequently forecasts clinical outcomes. By positioning severity as a basis for clinical outcomes, the two objectives mutually enhance each other.
Our experimental results robustly demonstrate the effectiveness of our method, indicating a promising future for LLMs in predictive tasks related to patient care.
Specifically, our model exhibits a notable enhancement in both Accuracy (ACC) and F1-score metrics for severity assessment and clinical outcome prediction tasks, when juxtaposed against conventional models.
In the task of predicting clinical outcomes, the traditional model struggles to accurately predict class 1 (death), whereas our model excels in this aspect, performing significantly better at predicting it.

\subsection{Findings in features}
Our study also employed the GradientBoost model for feature selection regarding disease severity and clinical outcomes, identifying the top five features associated with them, respectively.
The features related to clinical outcomes include Lymphocyte Percentage (LYMPH\%), hypersensitive C-reactive protein (hs-CRP), Neutrophil Percentage (Neu\%), Hypertension (HBP), and Age, while the features associated with disease severity are D-Dimer, LYMPH\%, Creatinine (Cre), Albumin (ALB), and Direct Bilirubin (BC). Among these two sets of features, LYMPH\% exhibited the most significant correlation, demonstrating important associations with both clinical outcomes and disease severity.
Our findings are consistent with prior research, indicating that LYMPH\% holds high importance in both contexts. As a critical indicator of immune system status, a reduction in Lymphocyte Percentage (LYMPH\%) typically signals adverse clinical outcomes \cite{zhang2020lymphocyte, tan2020lymphopenia, national2020new}. 
D-dimer levels are regarded as effective predictive factors for disease severity, closely linked to clinical deterioration and poor prognosis. 
Furthermore, these results align with previous literature, emphasizing the significance of D-dimer and hypersensitive C-reactive protein (hs-CRP) in assessing disease progression \cite{rostami2020d, yu2020d, ahirwar2022study}. Studies have shown that increased age may influence certain physiological parameters, thereby affecting disease severity and final outcomes \cite{davies2020age}. 
Neutrophil Percentage (Neu\%), as a marker of inflammatory activity, is significantly meaningful in predicting patient clinical outcomes \cite{peng2020neutrophil}. Hypertension is considered one of the most common comorbidities among patients infected with COVID-19, greatly increasing hospitalization and mortality risks \cite{peng2021role}. Additionally, variations in Creatinine (Cre), Albumin (ALB), and Direct Bilirubin (BC) levels are also crucial for understanding the prognosis of COVID-19, becoming promising biomarkers for assessing disease severity and prognosis \cite{alfano2021twenty, huang2020decreased, efat2022blood}.

\subsection{Strengths and Weakness}
In our study, we utilize a large language model (LLM) as the core prediction tool. Unlike traditional methods, our approach formats data as prompts, allowing for greater flexibility and enabling us to directly inform the model of any missing feature values. 
We view this as a specialized form of text-assisted prediction. 
Previous research has already suggested that combining the auxiliary text information will improve the performance of LLMs.
Our multi-objective learning strategy provides the model with more accurate labels for learning and effectively excludes instances of "mild \&\& death" as shown in Figure \ref{pic3}. 
We analyzed the outputs of CovidLLM and found no samples predicted as "mild \&\& death," which aligns with the absence of such cases in our dataset, underscoring the effectiveness of our strategy.
Regarding our dataset, the average patient age is 61, and it includes comprehensive historical records of conditions like diabetes and hypertension. 
This focus on older patients and individuals with comorbidities enhances the relevance of our findings. 
However, our study does have limitations. We have not fully leveraged the advantages of the LLMs by incorporating patient text information and other auxiliary data. Additionally, all patients in our dataset were unvaccinated.

\subsection{Future Works}
Given the flexible input format of LLMs, we plan to incorporate additional information for predicting the severity and clinical outcomes of COVID-19 patients, including patient self-reports and other textual data.
LLMs demonstrate strong predictive capabilities in time series, and the relationship between certain serological indicators and disease severity can change over time \cite{bib37, bib38, bib39}. Therefore, we aim to integrate time series data and serological indicators as inputs for predicting severity and clinical outcomes.
Furthermore, we intend to explore the effectiveness and potential of LLMs in predicting severity and clinical outcomes for other diseases, broadening the application of our findings.

\section{Conclusion}
Our research uses the GradientBoost algorithm to screen features related to disease severity and clinical outcomes and then designs these features as special Prompts. 
Use these special prompt and multi-objective learning strategies to fine-tune large language models to predict the severity and clinical outcomes of patients.
The features related to severity screened out in our study include D-Dimer, Lymphocyte Percentage (LYMPH\%), Creatinine (Cre), Albumin (ALB), and Direct Bilirubin (BC).
The characteristics related to clinical outcomes screened out in our study include Lymphocyte Percentage (LYMPH\%), hypersensitive C-reactive protein (hs-CRP), Neutrophil Percentage (Neu\%), Hypertension (HBP), and Age.
Our comparative experiments show that our model has better performance compared to traditional models. Our model achieves the highest accuracy in the task of predicting disease severity and predicting clinical outcomes.
Additionally, in clinic outcome prediction tasks, our model has a significant improvement in class 1 (death). The performance in predicting class 1 (death) is more important in predicting class 0 (survive).

\bibliographystyle{ieeetr} 
\bibliography{reference.bib} 

\begin{thebibliography}{10}

\bibitem{fernandez2021prevalence}
C.~Fern{\'a}ndez-de Las-Pe{\~n}as, D.~Palacios-Ce{\~n}a, V.~G{\'o}mez-Mayordomo, L.~L. Florencio, M.~L. Cuadrado, G.~Plaza-Manzano, and M.~Navarro-Santana, ``Prevalence of post-covid-19 symptoms in hospitalized and non-hospitalized covid-19 survivors: A systematic review and meta-analysis,'' {\em European journal of internal medicine}, vol.~92, pp.~55--70, 2021.

\bibitem{weng2021pain}
L.-M. Weng, X.~Su, and X.-Q. Wang, ``Pain symptoms in patients with coronavirus disease (covid-19): a literature review,'' {\em Journal of Pain Research}, pp.~147--159, 2021.

\bibitem{fernandez2021headache}
C.~Fern{\'a}ndez-de-las Pe{\~n}as, M.~Navarro-Santana, V.~G{\'o}mez-Mayordomo, M.~L. Cuadrado, D.~Garc{\'\i}a-Azor{\'\i}n, L.~Arendt-Nielsen, and G.~Plaza-Manzano, ``Headache as an acute and post-covid-19 symptom in covid-19 survivors: A meta-analysis of the current literature,'' {\em European journal of neurology}, vol.~28, no.~11, pp.~3820--3825, 2021.

\bibitem{cirulli2020long}
E.~T. Cirulli, K.~M. Schiabor~Barrett, S.~Riffle, A.~Bolze, I.~Neveux, S.~Dabe, J.~J. Grzymski, J.~T. Lu, and N.~L. Washington, ``Long-term covid-19 symptoms in a large unselected population,'' {\em medrxiv}, pp.~2020--10, 2020.

\bibitem{mahase2021covid}
E.~Mahase, ``Covid-19: Sore throat, fatigue, and myalgia are more common with new uk variant,'' 2021.

\bibitem{kahraman2020mucosal}
F.~C. Kahraman and H.~{\c{C}}a{\c{s}}kurlu, ``Mucosal involvement in a covid-19-positive patient: a case report,'' {\em Dermatologic therapy}, vol.~33, no.~4, 2020.

\bibitem{berlin2020severe}
D.~A. Berlin, R.~M. Gulick, and F.~J. Martinez, ``Severe covid-19,'' {\em New England Journal of Medicine}, vol.~383, no.~25, pp.~2451--2460, 2020.

\bibitem{he2021clinical}
X.~He, X.~Cheng, X.~Feng, H.~Wan, S.~Chen, and M.~Xiong, ``Clinical symptom differences between mild and severe covid-19 patients in china: a meta-analysis,'' {\em Frontiers in public health}, vol.~8, p.~561264, 2021.

\bibitem{hu2021cytokine}
B.~Hu, S.~Huang, and L.~Yin, ``The cytokine storm and covid-19,'' {\em Journal of medical virology}, vol.~93, no.~1, pp.~250--256, 2021.

\bibitem{alballa2021machine}
N.~Alballa and I.~Al-Turaiki, ``Machine learning approaches in covid-19 diagnosis, mortality, and severity risk prediction: A review,'' {\em Informatics in medicine unlocked}, vol.~24, p.~100564, 2021.

\bibitem{chieregato2022hybrid}
M.~Chieregato, F.~Frangiamore, M.~Morassi, C.~Baresi, S.~Nici, C.~Bassetti, C.~Bn{\`a}, and M.~Galelli, ``A hybrid machine learning/deep learning covid-19 severity predictive model from ct images and clinical data,'' {\em Scientific reports}, vol.~12, no.~1, p.~4329, 2022.

\bibitem{wu2021interpretable}
H.~Wu, W.~Ruan, J.~Wang, D.~Zheng, B.~Liu, Y.~Geng, X.~Chai, J.~Chen, K.~Li, S.~Li, {\em et~al.}, ``Interpretable machine learning for covid-19: An empirical study on severity prediction task,'' {\em IEEE Transactions on Artificial Intelligence}, vol.~4, no.~4, pp.~764--777, 2021.

\bibitem{hu2020early}
C.~Hu, Z.~Liu, Y.~Jiang, O.~Shi, X.~Zhang, K.~Xu, C.~Suo, Q.~Wang, Y.~Song, K.~Yu, {\em et~al.}, ``Early prediction of mortality risk among patients with severe covid-19, using machine learning,'' {\em International journal of epidemiology}, vol.~49, no.~6, pp.~1918--1929, 2020.

\bibitem{patterson2021immune}
B.~K. Patterson, J.~Guevara-Coto, R.~Yogendra, E.~B. Francisco, E.~Long, A.~Pise, H.~Rodrigues, P.~Parikh, J.~Mora, and R.~A. Mora-Rodr{\'\i}guez, ``Immune-based prediction of covid-19 severity and chronicity decoded using machine learning,'' {\em Frontiers in immunology}, vol.~12, p.~700782, 2021.

\bibitem{sayed2021applying}
S.~A.-F. Sayed, A.~M. Elkorany, and S.~S. Mohammad, ``Applying different machine learning techniques for prediction of covid-19 severity,'' {\em Ieee Access}, vol.~9, pp.~135697--135707, 2021.

\bibitem{raman2023machine}
G.~Raman, B.~Ashraf, Y.~K. Demir, C.~D. Kershaw, S.~Cheruku, M.~Atis, A.~Atis, M.~Atar, W.~Chen, I.~Ibrahim, {\em et~al.}, ``Machine learning prediction for covid-19 disease severity at hospital admission,'' {\em BMC Medical Informatics and Decision Making}, vol.~23, no.~1, p.~46, 2023.

\bibitem{huang2023chatgpt}
H.~Huang, O.~Zheng, D.~Wang, J.~Yin, Z.~Wang, S.~Ding, H.~Yin, C.~Xu, R.~Yang, Q.~Zheng, {\em et~al.}, ``Chatgpt for shaping the future of dentistry: the potential of multi-modal large language model,'' {\em International Journal of Oral Science}, vol.~15, no.~1, p.~29, 2023.

\bibitem{xue2023promptcast}
H.~Xue and F.~D. Salim, ``Promptcast: A new prompt-based learning paradigm for time series forecasting,'' {\em IEEE Transactions on Knowledge and Data Engineering}, 2023.

\bibitem{liang2024exploring}
Y.~Liang, Y.~Liu, X.~Wang, and Z.~Zhao, ``Exploring large language models for human mobility prediction under public events,'' {\em Computers, Environment and Urban Systems}, vol.~112, p.~102153, 2024.

\bibitem{ding2024semantic}
Z.~Ding, J.~Tian, Z.~Wang, J.~Zhao, and S.~Li, ``Semantic understanding and data imputation using large language model to accelerate recommendation system,'' {\em arXiv preprint arXiv:2407.10078}, 2024.

\bibitem{jin2023time}
M.~Jin, S.~Wang, L.~Ma, Z.~Chu, J.~Y. Zhang, X.~Shi, P.-Y. Chen, Y.~Liang, Y.-F. Li, S.~Pan, {\em et~al.}, ``Time-llm: Time series forecasting by reprogramming large language models,'' {\em arXiv preprint arXiv:2310.01728}, 2023.

\bibitem{glm2024chatglm}
T.~GLM, A.~Zeng, B.~Xu, B.~Wang, C.~Zhang, D.~Yin, D.~Rojas, G.~Feng, H.~Zhao, H.~Lai, {\em et~al.}, ``Chatglm: A family of large language models from glm-130b to glm-4 all tools,'' {\em arXiv preprint arXiv:2406.12793}, 2024.

\bibitem{liu2022p}
X.~Liu, K.~Ji, Y.~Fu, W.~Tam, Z.~Du, Z.~Yang, and J.~Tang, ``P-tuning: Prompt tuning can be comparable to fine-tuning across scales and tasks,'' in {\em Proceedings of the 60th Annual Meeting of the Association for Computational Linguistics (Volume 2: Short Papers)}, pp.~61--68, 2022.

\bibitem{AdamBoost}
Y.~Freund and R.~E. Schapire, ``A desicion-theoretic generalization of on-line learning and an application to boosting,'' in {\em European conference on computational learning theory}, pp.~23--37, Springer, 1995.

\bibitem{gradientboost}
J.~H. Friedman, ``Greedy function approximation: a gradient boosting machine,'' {\em Annals of statistics}, pp.~1189--1232, 2001.

\bibitem{ho1998random}
T.~K. Ho, ``The random subspace method for constructing decision forests,'' {\em IEEE transactions on pattern analysis and machine intelligence}, vol.~20, no.~8, pp.~832--844, 1998.

\bibitem{peterson2009k}
L.~E. Peterson, ``K-nearest neighbor,'' {\em Scholarpedia}, vol.~4, no.~2, p.~1883, 2009.

\bibitem{zhang2020lymphocyte}
W.~Zhang, Z.~Zhang, Y.~Ye, Y.~Luo, S.~Pan, H.~Qi, Z.~Yu, and J.~Qu, ``Lymphocyte percentage and hemoglobin as a joint parameter for the prediction of severe and nonsevere covid-19: a preliminary study,'' {\em Annals of Translational Medicine}, vol.~8, no.~19, 2020.

\bibitem{tan2020lymphopenia}
L.~Tan, Q.~Wang, D.~Zhang, J.~Ding, Q.~Huang, Y.-Q. Tang, Q.~Wang, and H.~Miao, ``Lymphopenia predicts disease severity of covid-19: a descriptive and predictive study,'' {\em Signal transduction and targeted therapy}, vol.~5, no.~1, p.~33, 2020.

\bibitem{national2020new}
N.~H. Commission {\em et~al.}, ``New coronavirus pneumonia diagnosis and treatment plan (trial version 7)[r],'' {\em Beijing: National Health Commission}, 2020.

\bibitem{rostami2020d}
M.~Rostami and H.~Mansouritorghabeh, ``D-dimer level in covid-19 infection: a systematic review,'' {\em Expert review of hematology}, vol.~13, no.~11, pp.~1265--1275, 2020.

\bibitem{yu2020d}
H.-H. Yu, C.~Qin, M.~Chen, W.~Wang, and D.-S. Tian, ``D-dimer level is associated with the severity of covid-19,'' {\em Thrombosis research}, vol.~195, pp.~219--225, 2020.

\bibitem{ahirwar2022study}
A.~K. Ahirwar, R.~Takhelmayum, A.~Sakarde, B.~D. Rathod, P.~K. Jha, R.~Kumawat, and N.~Gopal, ``The study of serum hscrp, ferritin, il-6 and plasma d-dimer in covid-19: a retrospective study,'' {\em Hormone Molecular Biology and Clinical Investigation}, vol.~43, no.~3, pp.~337--344, 2022.

\bibitem{davies2020age}
N.~G. Davies, P.~Klepac, Y.~Liu, K.~Prem, M.~Jit, and R.~M. Eggo, ``Age-dependent effects in the transmission and control of covid-19 epidemics,'' {\em Nature medicine}, vol.~26, no.~8, pp.~1205--1211, 2020.

\bibitem{peng2020neutrophil}
F.~Peng, S.~Lei, C.~Wu, B.~Yu, Y.~Zhong, and S.~Wu, ``Neutrophil percentage and neutrophil-to-monocyte ratio as independent risk factors in the severity of covid-19,'' 2020.

\bibitem{peng2021role}
M.~Peng, J.~He, Y.~Xue, X.~Yang, S.~Liu, and Z.~Gong, ``Role of hypertension on the severity of covid-19: A review,'' {\em Journal of cardiovascular pharmacology}, vol.~78, no.~5, pp.~e648--e655, 2021.

\bibitem{alfano2021twenty}
G.~Alfano, A.~Ferrari, F.~Fontana, G.~Mori, G.~Ligabue, S.~Giovanella, R.~Magistroni, M.~Meschiari, E.~Franceschini, M.~Menozzi, {\em et~al.}, ``Twenty-four-hour serum creatinine variation is associated with poor outcome in the novel coronavirus disease 2019 (covid-19) patients,'' {\em Kidney Research and Clinical Practice}, vol.~40, no.~2, p.~231, 2021.

\bibitem{huang2020decreased}
W.~Huang, C.~Li, Z.~Wang, H.~Wang, N.~Zhou, J.~Jiang, L.~Ni, X.~A. Zhang, and D.-W. Wang, ``Decreased serum albumin level indicates poor prognosis of covid-19 patients: hepatic injury analysis from 2,623 hospitalized cases,'' {\em Science China Life Sciences}, vol.~63, pp.~1678--1687, 2020.

\bibitem{efat2022blood}
A.~Efat, S.~Shoeib, A.~ElKholy, O.~S. Hussein~Aboelela, and D.~Elshamy, ``Blood phenotype o and indirect bilirubin are associated with lower, early covid-19—related mortality: A retrospective study,'' {\em International Journal of Immunopathology and Pharmacology}, vol.~36, p.~03946320221133952, 2022.

\bibitem{bib37}
B.~Sun, Y.~Feng, X.~Mo, P.~Zheng, and L.~Chen, ``Kinetics of sars-cov-2 specific igm and igg responses in covid-19 patients,'' {\em Emerging Microbes and Infections}, vol.~9, no.~1126, pp.~1--36, 2020.

\bibitem{bib38}
N.~Post, D.~Eddy, C.~Huntley, M.~C. I.~V. Schalkwyk, M.~Shrotri, D.~Leeman, S.~Rigby, S.~V. Williams, W.~H. Bermingham, and P.~a. Kellam, ``Antibody response to sars-cov-2 infection in humans: A systematic review,'' {\em PLOS ONE}, vol.~15, 2020.

\bibitem{bib39}
J.~V. Elslande, E.~Houben, M.~Depypere, A.~Brackenier, S.~Desmet, E.~André, M.~V. Ranst, K.~Lagrou, and P.~Vermeersch, ``Diagnostic performance of seven rapid igg/igm antibody tests and the euroimmun iga/igg elisa in covid-19 patients,'' {\em Clinical Microbiology and Infection}, vol.~26, no.~8, pp.~1082--1087, 2020.

\end{thebibliography}
\end{document}